\documentclass{bmvc2k}
\usepackage[section]{placeins}
\usepackage{amssymb}
\usepackage{makecell}
\usepackage{bbm}
\usepackage{amsmath}
\usepackage{flafter}       
\usepackage{dblfloatfix}   
\usepackage{wrapfig}
\usepackage{afterpage}

\usepackage[inline]{enumitem}

\title{Open-Vocabulary Audio-Visual Event Localization via Complex-Valued Fusion}

\addauthor{Anirudh Praveen}{anirudhp24@iitk.ac.in}{1}
\addauthor{Koteswar Rao Jerripothula}{kotesrj@iitk.ac.in}{1}
\addauthor{Pratik Joshi}{pratik.joshi@dolby.com}{2}
\addauthor{Aveen Dayal}{aveen.dayal@dolby.com}{2}
\addauthor{Neela Sawant}{neela.sawant@dolby.com}{2}

\addinstitution{
 IIT Kanpur\\
 Kanpur, India
}
\addinstitution{
 Dolby Laboratories\\
 Bengaluru, India
}

\runninghead{Praveen et al.}{Complex-Valued AV Fusion for OV-AVEL}

\def\etal{\emph{et al}\bmvaOneDot}

\begin{document}

\maketitle

\begin{abstract}

Open-Vocabulary Audio-Visual Event Localization (OV-AVEL) labels
each video segment with an event class, including classes that were
never seen during training. The dominant pipeline uses a frozen
multimodal foundation model (e.g.\ ImageBind) to embed the visual
frame, the audio mel-spectrogram, and each candidate class name into
a shared space, then computes two cosine similarities for each segment against each class: visual--text and audio--text. Existing methods then collapse
this pair into a single scalar score with a fixed rule (geometric
mean, weighted average) before taking the argmax. 
Instead, we compute complex-valued similarities and learn their fusion using a complex-valued neural network (CVNN). Each modality's standard representation becomes the
real part of our pipeline, and a paired companion stream
supplies the imaginary part. We use imaginary part of iHSV for visual modality and CycleGAN-translated phase spectrogram for audio modality as these companion streams. This results in two complex
similarities, which are then fused. While the vision and audio encoders
remain frozen, only the temporal-attention blocks and the fusion CVNN are trained.
The four-stream complex architecture sets a new state of the art on
both OV-AVEL benchmarks. On the \emph{open} (unseen-class) split of
OV-AVEBench we reach $\mathbf{66.5/59.1/54.1\%}$ Acc/Seg-F1/Event-F1
($\mathbf{+1.6/+4.1/+6.6}$ over the previously reported fine-tuned
baseline), with consistent gains for seen classes as well.
We also modify AVE dataset for this task and observe that our architecture reaches
$\mathbf{60.7/51.9/50.4\%}$ Acc/Seg-F1/Event-F1, achieving state-of-the-art OV-AVEL results on it as well. We also propose a two-stream alternative, which also sees great improvements over the baseline. Code is available at \url{https://github.com/AnirudhPraveen/Complex-Valued-OV-AVEL}

\end{abstract}

\section{Introduction}
\label{sec:intro}

Audio-visual event localization (AVEL) labels each video second with the event class that is simultaneously audible and visible~\cite{tian2018avel}. The closed-vocabulary formulation has been extensively studied via cross-modal fusion: attention-based networks~\cite{DAM,DMRN,CMRAN}, positive-sample propagation~\cite{PSP}, background suppression~\cite{CMBS}, and contrastive variants~\cite{ContrastiveLabel,EventSpecificAVEL,AVECLIP,CACE}. A related thread targets generalized audio-visual zero-shot learning for whole-video classification into unseen categories~\cite{CJME,AVGZSLNet,AVCA,TCAF,AVGZSLEasyWay}. Recent work pushes AVEL into open-set settings~\cite{OpenAVE}; the strict \emph{open-vocabulary} variant of Zhou~\etal~\cite{Zhou_2025_CVPR} goes further: test classes are never seen during training, and the model must produce segment-level labels by reasoning over text-defined categories at inference. The prevailing recipe uses a frozen audio-visual-text foundation model (ImageBind~\cite{girdhar2023imagebind}, LanguageBind~\cite{LanguageBind}, or CLIP/CLAP~\cite{CLIP,CLAP}) to embed audio, video, and text into a shared space, then predicts the class with the highest combined cosine similarity—\emph{open-vocabulary by construction}.

The bottleneck is fusion. For each segment and class $c$, frozen encoders yield two numbers: $\mathrm{sim}_{vt}[c]$ and $\mathrm{sim}_{at}[c]$. Existing OV-AVEL pipelines combine these with a fixed scalar rule (weighted average or geometric mean)~\cite{Zhou_2025_CVPR}, collapsing the pair to a single value and discarding a degree of freedom that could otherwise vary across classes and segments.

\begin{wrapfigure}{l}{0.48\textwidth}
\centering
\includegraphics[width=\linewidth, keepaspectratio]{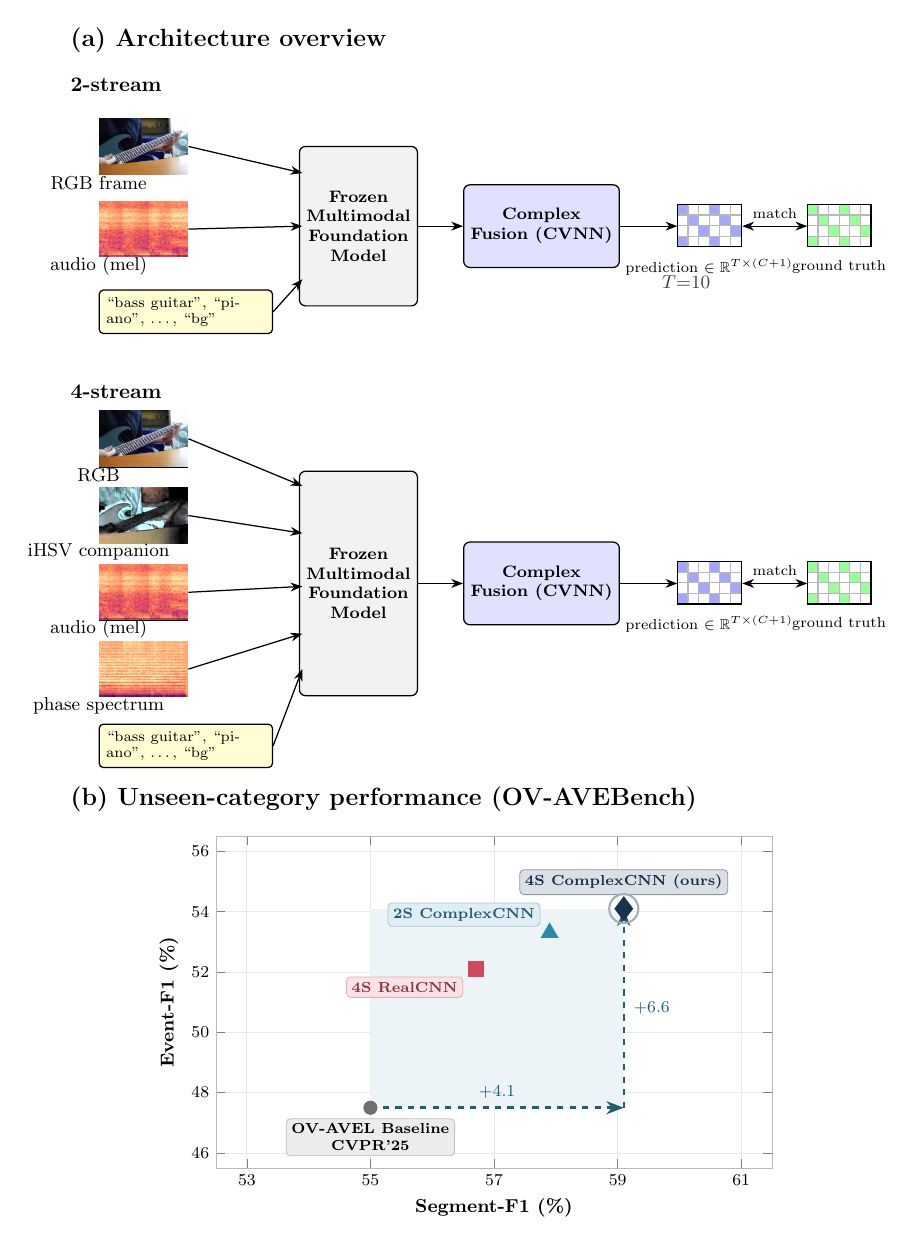}
\caption{\textbf{(a)} 2- and 4-stream complex-fusion pipelines.
  \textbf{(b)} Unseen-split Segment-F1 vs Event-F1 on OV-AVEBench;
  the 4-stream complex model improves the baseline by $+4.1$/$+6.6$.}
\label{fig:teaser}
\end{wrapfigure}

A related limitation: existing pipelines discard \emph{audio phase}. The standard ImageBind encoder consumes a log-mel spectrogram, discarding STFT phase—yet phase is a free by-product of the same Fourier transform and carries onset, harmonic, and transient cues that magnitude alone does not. We therefore (i) treat the visual--audio similarity pair as a complex value rather than collapsing it, and (ii) feed audio phase as a complementary, mel-shaped input to the same frozen encoder.
\vspace{-1.5em}
\paragraph{Our approach.}
Our architecture has two testable ingredients (Fig.~\ref{fig:teaser}; schematic in Fig.~\ref{fig:architecture}; details in Sec.~\ref{sec:method}).

\textbf{(1) Complex-valued fusion:} We replace the scalar geometric-mean fusion of~\cite{Zhou_2025_CVPR} with a five-layer \emph{ComplexCNN}~\cite{Trabelsi2018Deep} that treats visual--text and audio--text similarities as the real and imaginary parts of a complex input. This 2-stream fusion accounts for the bulk of the gain over the baseline; a parameter-matched real CNN at the same stream count actually \emph{trails} it, confirming the lift stems from the complex parameterisation rather than extra capacity (Tab.~\ref{tab:complex-vs-real}).

\textbf{(2) Four-stream extension:} Each modality is enriched with a paired companion—an \emph{iHSV}~\cite{FCCNs} transform on the visual side and a \emph{CycleGAN-translated phase} spectrogram on the audio side—giving each modality its own complex (content $+$ structure) pair, both fed to the same ComplexCNN. The auxiliary streams add a smaller, complementary gain on top of (1) (Tab.~\ref{tab:complex-vs-real}).
\vspace{-1.5em}
\paragraph{Contributions:}
\begin{enumerate*}
\item A \emph{complex-valued fusion network} for OV-AVEL treating audio--text similarity as the imaginary part and visual--text similarity as the real part, fused by a five-layer ComplexCNN. This 2-stream design yields the dominant gain over~\cite{Zhou_2025_CVPR} (Tab.~\ref{tab:complex-vs-real}, Sec.~\ref{sec:method}).
\item A \emph{four-stream extension} enriching each modality with a paired imaginary companion (iHSV visual, CycleGAN-phase audio), stacked as two complex channels into the same ComplexCNN, adding a further complementary gain (Tab.~\ref{tab:complex-vs-real}; Sec.~\ref{sec:method-vt}--\ref{sec:method-audio-imag}).
\end{enumerate*}

{Our core motivation is to explore complex-valued neural networks (CVNNs) or ComplexCNNs for multimodal fusion. CVNNs yield richer representations than real-valued networks, provided complex inputs are supplied. Two natural ways exist with audio-video data: (i) a separate complex channel per modality, or (ii) combining both modalities into one complex channel. The 4-stream approach uses option (i), yielding two complex-valued input channels. The 2-stream approach uses option (ii), packaging both modalities into a single complex channel—real for visual, imaginary for audio—giving a simpler, more efficient, and redundancy-reduced representation. The intuition is that audio information is directionally orthogonal to visual information, analogous to depth.}
 \section{Related Work}
  \label{sec:related}

  \textbf{Closed-set audio-visual event localization:}
  Tian~\etal~\cite{tian2018avel} introduced the AVE benchmark and
  formulated AVEL as a fully-supervised, closed-vocabulary, per-segment
  classification problem in which each one-second segment is labelled
  with a single audio-visual event category or background. Follow-up work
  has primarily strengthened the \emph{cross-modal fusion}: dual-modality
  attention modules (DAM~\cite{DAM}, DMRN~\cite{DMRN}),
  relation-aware cross-modal attention (CMRAN~\cite{CMRAN}),
  positive-sample propagation (PSP~\cite{PSP}), cross-modal background
  suppression (CMBS~\cite{CMBS}), contrastive
  learning~\cite{ContrastiveLabel}, event-specific
  preferences~\cite{EventSpecificAVEL}, CLIP-based temporal
  transformers~\cite{AVECLIP}, and co-guidance
  attention~\cite{CACE}; a unified multi-task family also handles
  audio-visual video parsing (MM-Pyramid~\cite{MMPyramid}) and
  segmentation~\cite{AVS}. All are \emph{closed-vocabulary}: test
  categories must overlap with training categories, and the fusion step
  reduces audio and visual evidence to real-valued scalars or
  concatenations.
  \textbf{Audio-visual zero-shot learning and open-set AVE:}
  Generalized zero-shot learning (AV-GZSL)~\cite{CJME,AVGZSLNet} treats
  unseen classes at the \emph{video} level, classifying a whole clip into
  a possibly unseen category by aligning audio-visual features with text,
  more recently via cross-modal attention~\cite{AVCA,TCAF} and simpler
  distillation~\cite{AVGZSLEasyWay}. A closer setting is open-set AVE
  (\emph{OpenAVE}~\cite{OpenAVE}), which adds an ``unknown'' class for
  unseen categories but does not predict their specific labels. We address
  the strictly harder \emph{open-vocabulary} variant, producing explicit
  segment-level labels for unseen classes.
  \textbf{Open-vocabulary AVEL:}
  Zhou~\etal~\cite{Zhou_2025_CVPR} introduced OV-AVEL by reusing the
  ImageBind~\cite{girdhar2023imagebind} text encoder so the test
  vocabulary can include unseen classes. They report a \emph{training-free}
  baseline (predict an event only when audio and visual independently rank
  the same class top-1) and a \emph{fine-tuned} baseline (a small
  real-valued temporal-attention head and fusion CNN on frozen ImageBind
  features); both combine a pair of scalar text similarities with a fixed
  rule before the argmax. We keep the open-vocabulary protocol but replace
  the scalar fusion with a complex-valued fusion of the visual--text and
  audio--text similarities (Sec.~\ref{sec:intro}).
  \textbf{Multimodal foundation models:}
  Large-scale image--text contrastive pretraining (CLIP~\cite{CLIP}) and
  its multi-modal extensions (ImageBind~\cite{girdhar2023imagebind},
  LanguageBind~\cite{LanguageBind}, audio--text CLAP~\cite{CLAP}) provide a
  shared embedding space, enabling open-vocabulary inference by
  construction: a new class name is encoded into the shared text space and
  compared against modality embeddings via cosine similarity. We use
  ImageBind as a frozen backbone like the OV-AVEL
  baseline~\cite{Zhou_2025_CVPR}; our contribution is at the fusion rather
  than the representation stage.
  \textbf{Complex-valued neural networks:}
  Complex-valued networks~\cite{Trabelsi2018Deep} are a parameter-tied
  form of real network in which weights $W = W_R + iW_I$ act on a complex
  input $x = x_R + i x_I$ through structured cross-channel mixing
  $Wx = (W_R x_R - W_I x_I) + i(W_R x_I + W_I x_R)$. Used in MRI
  reconstruction, audio source separation, and other intrinsically complex
  signal-processing tasks, their use in open-vocabulary multimodal event
  localization is, to our knowledge, new.
  \textbf{Phase information in audio:}
  The STFT phase carries cues complementary to magnitude (onsets, harmonic
  locking, transient
  localization)~\cite{Paliwal2011PhaseImportance,Mowlaee2016PhaseAware} but
  is hard for a magnitude-trained encoder to consume because of $2\pi$
  wrap-around. A common fix is the instantaneous-frequency (IF) derivative,
  smooth across the wrap and locally concentrated on harmonics and
  onsets~\cite{Boashash1992IF,Engel2019GANSynth}; separately,
  CycleGAN-style unpaired translation has been applied to spectrogram-domain
  audio such as voice
  conversion~\cite{Kaneko2018CycleGANVC,Zhu2017Unpaired}. We combine these
  threads, not done before to our knowledge: an unpaired CycleGAN translates
  the IF spectrogram into the mel-magnitude manifold a frozen audio encoder
  expects, letting it ingest phase-derived input without retraining.
  \textbf{Color spaces and visual structure:}
  HSV decomposes a colour into a hue angle $H \in [0, 2\pi]$, saturation
  $S \in [0,1]$ and value $V \in [0,1]$, the angular $H$ being the natural
  visual analogue of audio phase. The synthetic \emph{iHSV} imaginary
  companion of an RGB frame (an involutive flip of the HSV decomposition)
  was introduced by Yadav and Jerripothula~\cite{FCCNs} as the imaginary
  input to a fully complex-valued image classifier; we adopt it as the
  imaginary half of the visual stream.

\section{Methodology}
\label{sec:method}

\begin{figure*}[t]
\centering
\includegraphics[width=\textwidth]{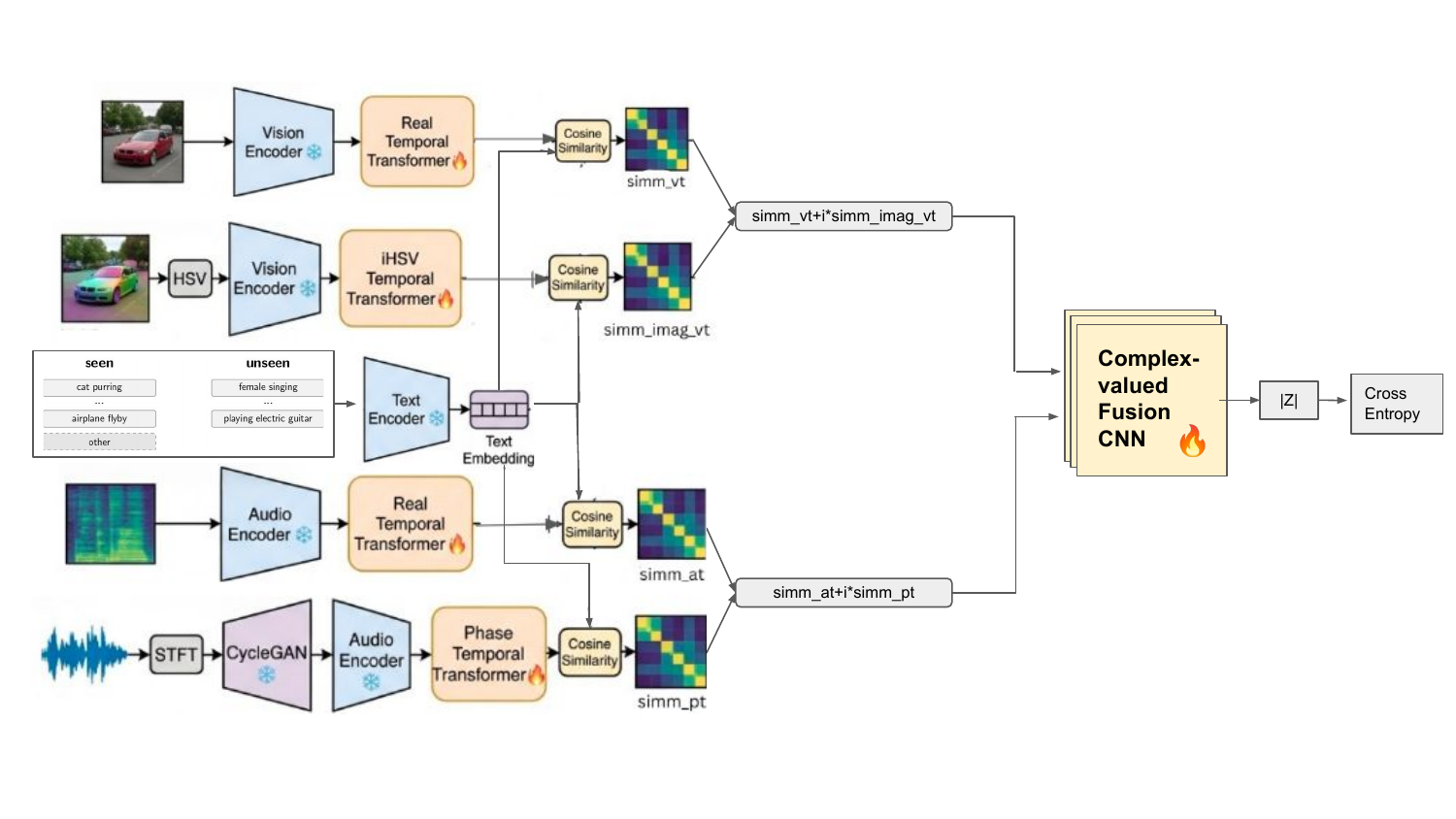}
\vspace{-4em}
\caption{\textbf{Four-stream complex-valued OV-AVEL architecture.}
  Each modality forms a complex similarity tensor against the text prompts:
  $\mathbf{Z}_v=\mathbf{S}_{vt}+i\,\mathbf{S}_{vt}^{\textsc{i}}$ (RGB
  $+i\cdot$ iHSV, via frozen $\mathcal{E}_v$) and
  $\mathbf{Z}_a=\mathbf{S}_{at}+i\,\mathbf{S}_{pt}$ (mel $+i\cdot$
  CycleGAN-translated phase, via frozen $\mathcal{E}_a$ and frozen
  $G_{\mathrm{IF}\to\mathrm{mel}}$). The two tensors are stacked as the two
  channels of a five-layer ComplexCNN whose output magnitude gives the
  per-segment class scores. Only the four temporal-attention heads
  ($\Phi_v,\Phi_v^{\textsc{i}},\Phi_a,\Phi_\phi$) and the ComplexCNN are
  trained; the ImageBind encoders and the CycleGAN generator are frozen}
\label{fig:architecture}
\vspace{-1.5em}
\end{figure*}

Our model takes a 10-second video sliced into $T{=}10$ one-second
segments and produces, for each segment, a label drawn from an unseen
category set $\mathcal{C}=\{c_1,\dots,c_C,\textsc{bg}\}$, where the
test categories may be unseen at training and $\textsc{bg}$ denotes
non-event background. We keep the open-vocabulary protocol of
Zhou~\etal~\cite{Zhou_2025_CVPR} (frozen foundation model,
text-aligned cosine similarities, class-agnostic fusion) but replace
the two-stream (RGB, mel) scalar pipeline with a \emph{
complex-valued} one. Sec.~\ref{sec:method-backbone} fixes notation;
Secs.~\ref{sec:method-vt}--\ref{sec:method-audio-imag} describe the
four streams; Sec.~\ref{sec:fusion} describes the complexcnn fusion;
Sec.~\ref{sec:method-train} specifies what is trainable.
Fig.~\ref{fig:architecture} gives a schematic overview.

\subsection{Frozen backbone and notation}
\label{sec:method-backbone}

Each video is represented by a log-mel spectrogram
$\mathbf{a}\in\mathbb{R}^{1\times 128 \times 204}$, $T$ RGB frames
$\mathbf{V}=\{\mathbf{v}_t\}_{t=1}^{T}$ with
$\mathbf{v}_t\in\mathbb{R}^{3\times 224\times 224}$, and a text prompt
set $\mathbf{T}=\{\tau_1,\dots,\tau_{C+1}\}$ that includes a literal
``Background'' prompt. We use ImageBind ViT-H~\cite{girdhar2023imagebind}
as a frozen multimodal backbone with encoders $\mathcal{E}_a$,
$\mathcal{E}_v$, $\mathcal{E}_t$ that map every modality into a shared
$D{=}1024$-dim space:
\begin{equation}
\mathbf{f}_a = \mathcal{E}_a(\mathbf{a})\in\mathbb{R}^{T\times D},\quad
\mathbf{f}_v = \mathcal{E}_v(\mathbf{V})\in\mathbb{R}^{T\times D},\quad
\mathbf{f}_t = \mathcal{E}_t(\mathbf{T})\in\mathbb{R}^{(C+1)\times D}.
\end{equation}
All three encoders remain frozen at training time. This is what gives
the system its open-vocabulary property: a previously unseen test
class is classified simply by appending its name to $\mathbf{T}$ at
inference.

\subsection{Visual stream as a complex pair: RGB \texorpdfstring{$+\,i\cdot$}{ + i} imaginary component of iHSV}
\label{sec:method-vt}

We treat the visual modality as a single complex object: its
\emph{real part} is the standard RGB content view, and its
\emph{imaginary part} is a colour-structure companion derived from the imaginary part of  iHSV~\cite{FCCNs}. Both views share the
\emph{same} frozen vision encoder $\mathcal{E}_v$, so no new backbone parameters are introduced.

\paragraph{Why a paired visual companion:}
The audio side of our architecture forms a complex pair naturally:
mel magnitude is the real part, and CycleGAN-translated audio phase
is the imaginary part (Sec.~\ref{sec:method-audio-imag}). For the
ComplexCNN fusion to receive a symmetric, two-complex-channel input,
the visual modality also needs a companion stream that plays the
role of the imaginary part.


\paragraph{Real part: RGB content.}
Per-segment features $\mathbf{f}_v = \mathcal{E}_v(\mathbf{V})$ are
refined by a one-layer temporal self-attention module $\Phi_v$ that
exchanges information across the $T$ time steps, giving
$\tilde{\mathbf{f}}_v = \Phi_v(\mathbf{f}_v)$. A cosine similarity to
the text features produces the real part of the visual channel:
\begin{equation}
\mathbf{S}_{vt}[t,c] = cossim\big(\tilde{\mathbf{f}}_v[t],\, \mathbf{f}_t[c]\big),
\qquad \mathbf{S}_{vt}\in\mathbb{R}^{T\times(C+1)}.
\end{equation}

\paragraph{Imaginary part: iHSV colour-structure companion.}
Each RGB frame is converted to HSV with $H\in[0,2\pi]$ and
$S,V\in[0,1]$. We form the imaginary triplet
\begin{equation}
\mathbf{i}(\mathbf{v}_t) = \big(V,\; S\sin H,\; S\big),
\end{equation}
the natural counterpart of the real triplet $(SH, S\cos H, V)$
used in fully-complex HSV decompositions. Treating this triplet as a
synthetic HSV image and converting it back to RGB yields a 3-channel
tensor in which the cyclic part of the hue is preserved through
$\sin H$ and the chromatic structure is preserved through $S$. We
normalise the resulting image with the ImageBind RGB statistics.
The same frozen $\mathcal{E}_v$
ingests this companion image, after which a separate temporal
attention $\Phi_v^{\textsc{i}}$ yields the imaginary part of the
visual channel:
\begin{equation}
\mathbf{S}_{vt}^{\textsc{i}}[t,c] =
cossim\big(\Phi_v^{\textsc{i}}\big(\mathcal{E}_v(\mathbf{i}(\mathbf{V}))\big)[t],\, \mathbf{f}_t[c]\big).
\end{equation}
The visual stream thus emits one complex similarity tensor
$\mathbf{Z}_v = \mathbf{S}_{vt} + i\mathbf{S}_{vt}^{\textsc{i}}
\in\mathbb{C}^{T\times(C+1)}$.

\subsection{Audio stream as a complex pair: mel \texorpdfstring{$+\,i\cdot$}{ + i} CycleGAN-phase}
\label{sec:method-audio-imag}

Symmetrically, we treat the audio modality as a single complex object:
\emph{real part} = log-mel magnitude content, \emph{imaginary part} =
phase information translated into the same mel-compatible domain so
the frozen audio encoder $\mathcal{E}_a$ can ingest it.

\paragraph{Real part: mel magnitude.}
A one-layer temporal attention $\Phi_a$ refines
$\mathbf{f}_a = \mathcal{E}_a(\mathbf{a})$ into
$\tilde{\mathbf{f}}_a$ and produces similarities
$\mathbf{S}_{at}[t,c] = cossim(\tilde{\mathbf{f}}_a[t],\, \mathbf{f}_t[c])$.

\subsubsection{Imaginary part: CycleGAN-translated phase.}
The STFT phase carries onsets and transient cues
complementary to magnitude, but is too noisy in its raw form to be
consumed by a magnitude-trained encoder. We address this in two steps.

\begin{figure*}[t]
\centering
\includegraphics[width=\textwidth]{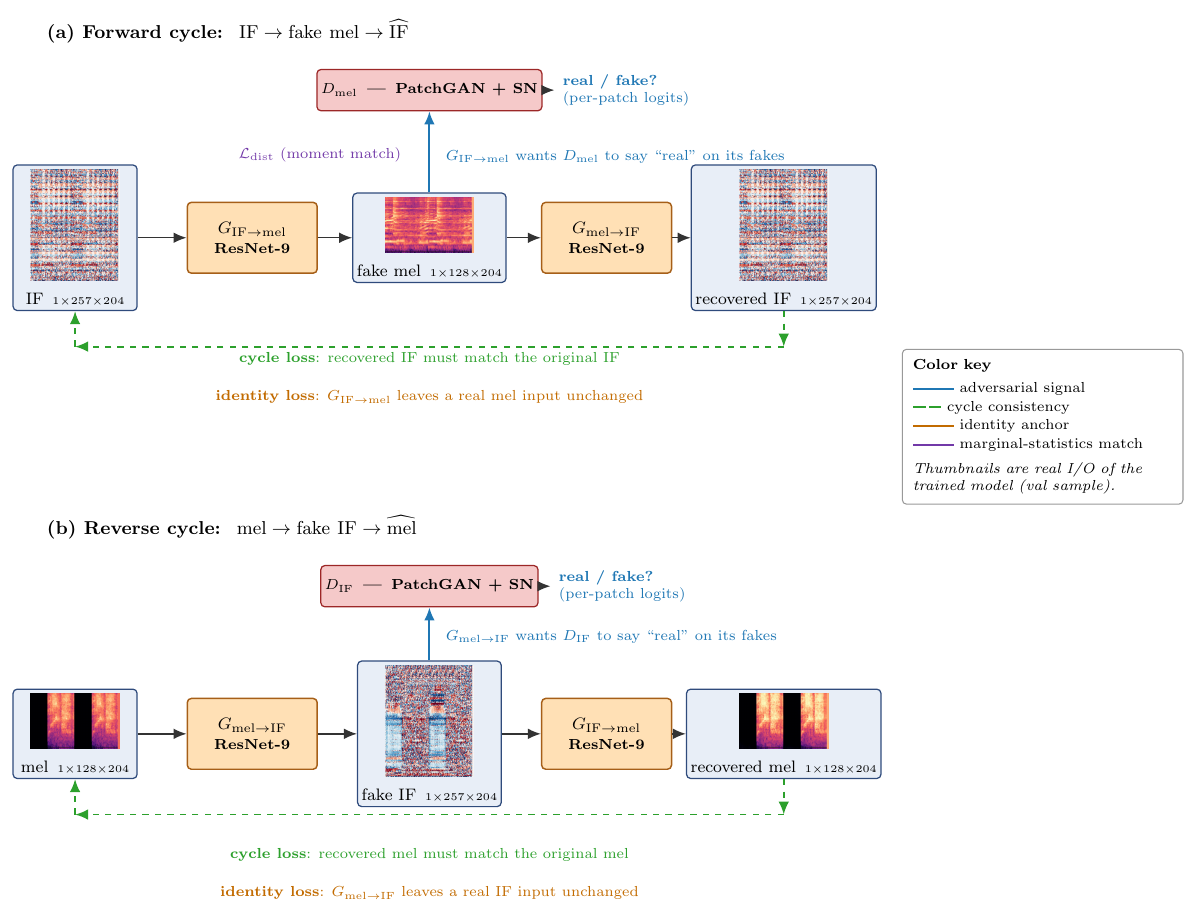}
  \caption{\textbf{Unpaired CycleGAN translating the instantaneous-frequency
  (IF) phase spectrogram into the mel-magnitude domain expected by the frozen
  audio encoder.} \emph{(a)} forward and \emph{(b)} reverse cycles, built from
  the two ResNet-9 generators $G_{\mathrm{IF}\to\mathrm{mel}}$,
  $G_{\mathrm{mel}\to\mathrm{IF}}$ and PatchGAN discriminators
  $D_{\mathrm{mel}},D_{\mathrm{IF}}$, trained with adversarial,
  cycle-consistency, identity, and distribution-matching
  ($\mathcal{L}_{\mathrm{dist}}$) losses. At OV-AVEL time only
  $G_{\mathrm{IF}\to\mathrm{mel}}$ is used; the rest is training-time scaffolding.}

\label{fig:cyclegan}
\vspace{-1.5em}
\end{figure*}

First, from the STFT phase
$\boldsymbol{\phi}\in[-\pi,\pi]^{1\times 257\times 204}$ we extract
the instantaneous frequency
$\mathrm{IF}[f,t] = (\boldsymbol{\phi}[f,t] - \boldsymbol{\phi}[f,t-1])
\bmod 2\pi - \pi$. Second, we train an unpaired
CycleGAN~\cite{Zhu2017Unpaired} with a generator
$G_{\mathrm{IF}\to\mathrm{mel}}:\mathbb{R}^{1\times 257\times 204}
\to\mathbb{R}^{1\times 128\times 204}$ and a companion
$G_{\mathrm{mel}\to\mathrm{IF}}$, plus two
PatchGAN~\cite{Isola2017pix2pix} discriminators
$D_{\mathrm{mel}}$ and $D_{\mathrm{IF}}$. The CycleGAN is
trained offline and \emph{frozen} before OV-AVEL training begins.
Fig.~\ref{fig:cyclegan} shows the two training cycles, the four loss
terms, and example translated spectrograms.

\vspace{-1mm}
\paragraph{Generator architecture:} Each generator is a $9$-residual-block ResNet in the style of Johnson
et al.~\cite{Johnson2016Perceptual}: the standard CycleGAN
backbone, with a $7{\times}7$ reflection-padded stem, two stride-$2$
downsampling convolutions, nine residual blocks, two
transpose-convolution upsamplers, and a $7{\times}7$ output convolution
with $\tanh$; instance norm and ReLU throughout.\\
\textbf{Training objectives:} We adopt the generic two-domain CycleGAN notation of
Zhu et al.~\cite{Zhu2017Unpaired}: domain $A$ is IF and domain $B$ is
mel, $a\!\sim\!A$ and $b\!\sim\!B$ are \emph{unpaired} samples (drawn
from different clips), $G\!\equiv\!G_{A\to B}\!=\!G_{\mathrm{IF}\to\mathrm{mel}}$,
its reverse $G_{B\to A}\!=\!G_{\mathrm{mel}\to\mathrm{IF}}$ plays the
role of $F$ in Zhu et al., and $D_B\!=\!D_{\mathrm{mel}}$ is the
discriminator for the codomain of $G$ (symmetrically $D_A\!=\!D_{\mathrm{IF}}$);
$\mathbb{E}_a,\mathbb{E}_b$ denote expectations over the two domains.
The generators are trained with four loss terms. The adversarial
loss uses the LSGAN~\cite{Mao2017LSGAN} formulation ($\ell_2$ on
discriminator outputs) for stability on spectrogram-shaped inputs:
\vspace{-1mm}
\begin{equation}
\mathcal{L}_{\mathrm{adv}}(G,D_B) =
\mathbb{E}_b\!\left[(D_B(b){-}1)^2\right] +
\mathbb{E}_a\!\left[D_B(G(a))^2\right],
\label{eq:adv}
\end{equation}
applied symmetrically for both directions. The cycle-consistency
loss preserves enough information through the round trip that the
input can be recovered:
\begin{equation}
\mathcal{L}_{\mathrm{cyc}} =
\mathbb{E}\!\left[\|G_{B\to A}(G_{A\to B}(a)) - a\|_1\right] +
\mathbb{E}\!\left[\|G_{A\to B}(G_{B\to A}(b)) - b\|_1\right].
\label{eq:cyc}
\end{equation}
The identity loss regularises each generator towards preserving its
target domain. In the standard CycleGAN this term feeds a real
target-domain sample directly to the generator, which is possible
because both domains share the same shape. Here the two domains have
different frequency dimensions, so $G_{\mathrm{IF}\to\mathrm{mel}}$ and
$G_{\mathrm{mel}\to\mathrm{IF}}$ cannot ingest a target-domain sample
unchanged; we therefore adaptive-pool the input to the generator's
expected input shape first. The term thus asks each generator to
recover the original target-domain sample from a resampled copy of
itself: a softer constraint than a strict no-op, but one that still
discourages spurious changes on in-domain inputs:
\begin{equation}
\mathcal{L}_{\mathrm{idt}} =
\mathbb{E}\!\left[\|G_{A\to B}(\mathrm{pool}_A(b)) - b\|_1\right] +
\mathbb{E}\!\left[\|G_{B\to A}(\mathrm{pool}_B(a)) - a\|_1\right].
\label{eq:idt}
\end{equation}

Finally, a distribution-matching loss on the IF$\to$mel direction
adds an explicit constraint that the marginal statistics of
$\hat{\mathbf{m}} = G_{\mathrm{IF}\to\mathrm{mel}}(\mathrm{IF})$
match those of real mel:
\begin{equation}
\mathcal{L}_{\mathrm{dist}} =
\|\mu_f(\hat{\mathbf{m}}){-}\mu_f(\mathbf{m})\|_1 +
\|\sigma_f(\hat{\mathbf{m}}){-}\sigma_f(\mathbf{m})\|_1 +
\tfrac{1}{2}\!\left(|\gamma{-}\hat{\gamma}|+|\kappa{-}\hat{\kappa}|\right),
\label{eq:dist}
\end{equation}
where $\mu_f,\sigma_f$ are per-frequency-bin mean/std and
$\gamma,\kappa$ are global skewness and kurtosis after
standardisation. The motivation for $\mathcal{L}_{\mathrm{dist}}$ is
specific to our frozen-backbone setting: the translated mel
$\hat{\mathbf{m}}$ is not an end in itself but is consumed by the
\emph{frozen} audio encoder $\mathcal{E}_a$, which cannot adapt to a
shifted input distribution. The adversarial term alone, with a
PatchGAN discriminator that judges only \emph{local} patch realism,
permits the global energy profile and dynamic range of
$\hat{\mathbf{m}}$ to drift away from the real mel manifold on which
$\mathcal{E}_a$ was pretrained; such drift pushes the frozen encoder
off-distribution and degrades the extracted features.
$\mathcal{L}_{\mathrm{dist}}$ anchors the per-bin marginal statistics
of $\hat{\mathbf{m}}$ to those of real mel so that $\mathcal{E}_a$
receives inputs in the distribution it expects. The total generator
objective combines all four:
\begin{equation}
\mathcal{L}_G = \mathcal{L}_{\mathrm{adv}} +
\lambda_{\mathrm{cyc}}\mathcal{L}_{\mathrm{cyc}} +
\lambda_{\mathrm{idt}}\mathcal{L}_{\mathrm{idt}} +
\lambda_{\mathrm{dist}}\mathcal{L}_{\mathrm{dist}},
\qquad
\lambda_{\mathrm{cyc}}{=}10,\;
\lambda_{\mathrm{idt}}{=}2,\;
\lambda_{\mathrm{dist}}{=}2.
\label{eq:total}
\end{equation}

\begin{wrapfigure}{r}{0.50\textwidth}

\centering
\includegraphics[width=\linewidth, keepaspectratio]{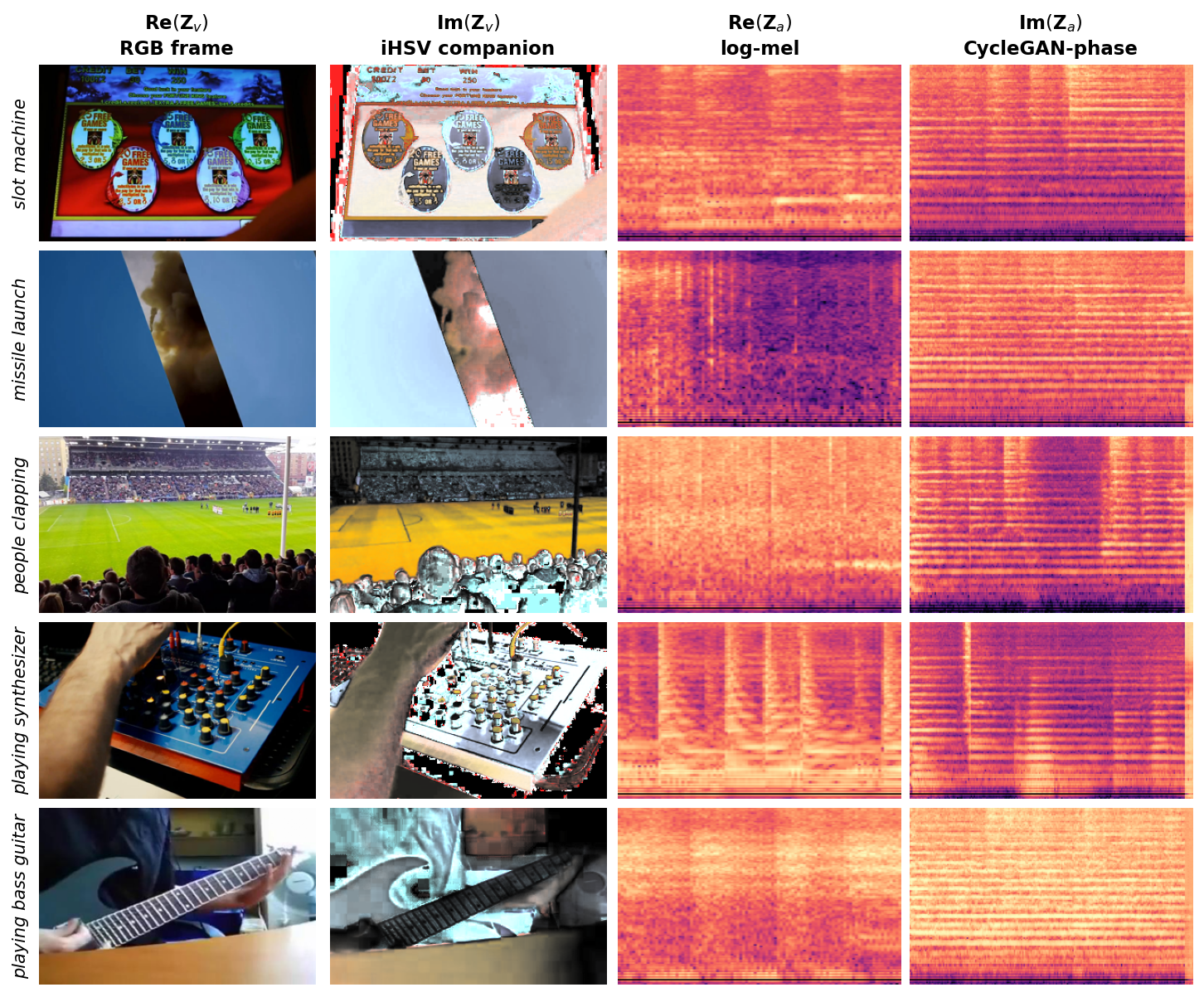}
\caption{\textbf{The four input streams forming our two complex
  channels}, shown for four event classes. 
  }

\label{fig:input-streams}

\end{wrapfigure}

The discriminators are PatchGAN~\cite{Isola2017pix2pix} networks
(5 conv layers, $4{\times}4$ kernels,
instance norm +
LeakyReLU($0.2$)), with spectral normalisation~\cite{Miyato2018SpectralNorm}
on every conv and instance noise ($\sigma{=}0.05$) added to the
discriminator input only during training. Adversarial training uses
one-sided label smoothing (real $= 0.9$), Adam with $\beta_1{=}0.5,\beta_2{=}0.999$, and a
cosine LR schedule; the discriminator update is skipped whenever
$\mathcal{L}_D{\le}0.1$, where $\mathcal{L}_D$ is the discriminator loss, to keep $D$ from dominating $G$.
The settings split into two groups. Values that match standard
CycleGAN practice~\cite{Zhu2017Unpaired} are kept at their canonical
defaults: cycle weight $\lambda_{\mathrm{cyc}}{=}10$, nine residual
blocks per generator, base learning rate $2{\times}10^{-4}$, and Adam
$(\beta_1,\beta_2){=}(0.5,0.999)$. The remaining values are set
\emph{empirically}, judged by the visual realism and per-frequency
moment match of the translated mel on a held-out split rather than by
any downstream metric: $\lambda_{\mathrm{idt}}{=}\lambda_{\mathrm{dist}}{=}2$,
a discriminator learning rate of $1{\times}10^{-4}$ (deliberately below the generator's, a two-timescale rule that stopped $D$ from
saturating early), batch size $16$, and $200$ epochs.

At OV-AVEL training time we use
$\hat{\mathbf{a}}_\phi = G_{\mathrm{IF}\to\mathrm{mel}}(\mathrm{IF})$,
$\mathbf{f}_\phi = \mathcal{E}_a(\hat{\mathbf{a}}_\phi)$, then a
temporal attention $\Phi_\phi$ to produce similarities
$\mathbf{S}_{pt}[t,c] = cossim(\Phi_\phi(\mathbf{f}_\phi)[t],\, \mathbf{f}_t[c])$.
The audio stream emits one complex similarity tensor
$\mathbf{Z}_a = \mathbf{S}_{at} + i\mathbf{S}_{pt}
\in\mathbb{C}^{T\times(C+1)}$.

\subsection{ComplexCNN fusion}
  \label{sec:method-fusion}
  \label{sec:fusion}

  \paragraph{2-stream packaging:} In the minimal variant we package the
  two scalar audio--visual similarities into a single complex channel
  \begin{equation}
  \mathbf{Z}_{\mathrm{2s}} \;=\; \mathbf{S}_{vt} \;+\; i\,\mathbf{S}_{at}
  \;\in\; \mathbb{C}^{1\times T\times(C+1)},
  \end{equation}
  so that real and imaginary parts carry the visual--text and
  audio--text similarities respectively. No companion streams are used.

\begin{wrapfigure}{r}{0.55\textwidth}

\centering
\includegraphics[width=\linewidth, keepaspectratio]{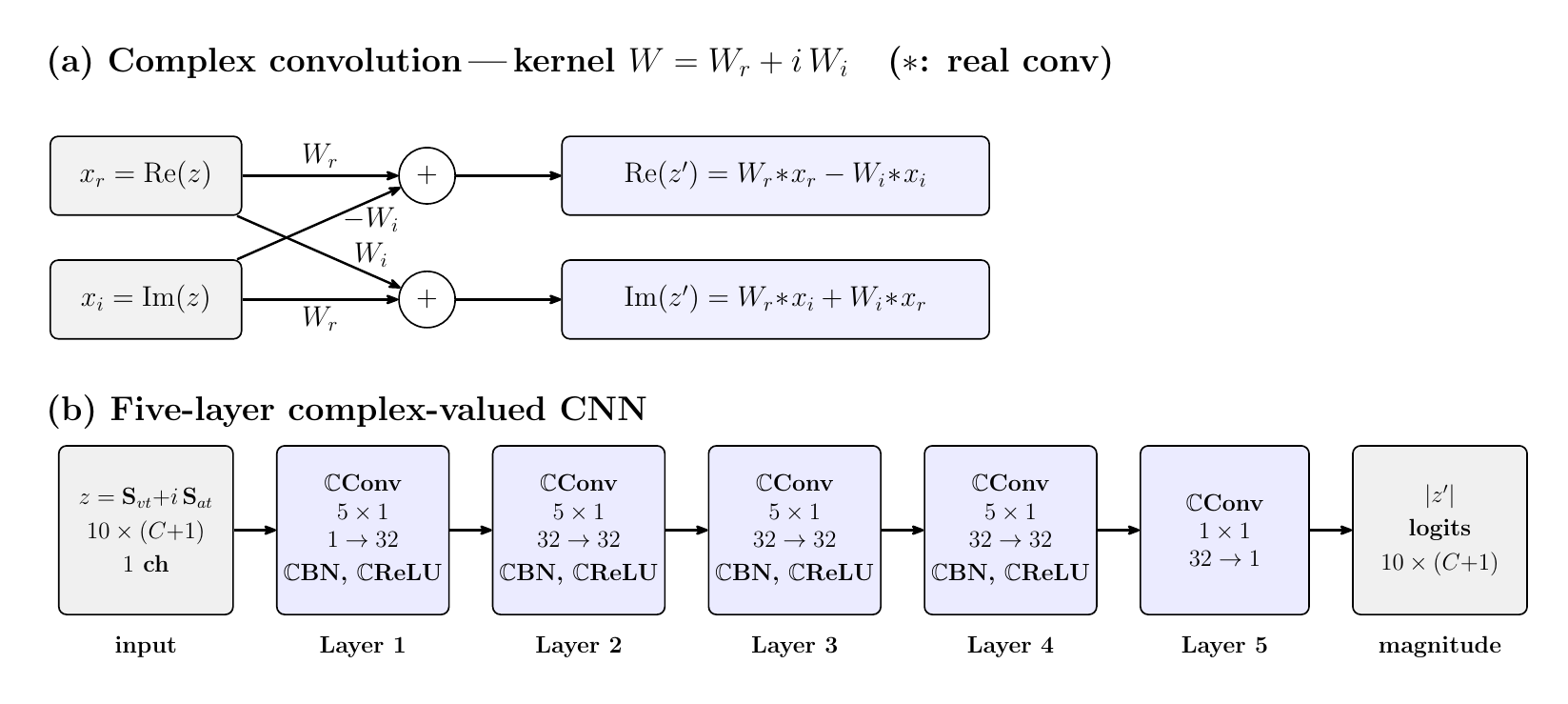}
\caption{\textbf{ComplexCNN fusion.}
    \textbf{(a)}~Complex conv: kernel $\mathbf{W}{=}\mathbf{W}_R{+}i\mathbf{W}_I$
    ties real/imaginary mixing as
    $\bigl[\begin{smallmatrix}\mathbf{W}_R&-\mathbf{W}_I\\
    \mathbf{W}_I&\mathbf{W}_R\end{smallmatrix}\bigr]$.
    \textbf{(b)}~Five layers: $4{\times}(5{\times}1)$ convs ($1{\to}32{\to}32$,
    BN+ReLU) $+$ $1{\times}1$ ($32{\to}1$); magnitude $\to$ scores.}
  \label{fig:complexcnn}

\end{wrapfigure}

  \paragraph{4-stream packaging:} In the full variant the visual complex
  tensor $\mathbf{Z}_v$ (Sec.~\ref{sec:method-vt}) and the audio complex
  tensor $\mathbf{Z}_a$ (Sec.~\ref{sec:method-audio-imag}) are stacked
  as the two channels of a complex-valued input
  $\mathbf{Z}\in\mathbb{C}^{2\times T\times(C+1)}$:
  \begin{equation}
  \mathbf{Z}^{(0)} = \mathbf{Z}_v = \mathbf{S}_{vt} + i\mathbf{S}_{vt}^{\textsc{i}},
  \qquad
  \mathbf{Z}^{(1)} = \mathbf{Z}_a = \mathbf{S}_{at} + i\mathbf{S}_{pt}.
  \label{eq:zpack}
  \end{equation}
  This per-modality pairing makes the audio--visual symmetry explicit:
  each complex channel describes one modality through a content +
  structure pair, and cross-modal interaction is delegated to the
  convolutional kernels rather than baked into the channel layout.
\vspace{-1em}
  \paragraph{Complex convolution:} The fusion network is a five-layer
    ComplexCNN. Write the input to a layer as a complex feature map
    $\mathbf{x}=\mathbf{x}_R + i\,\mathbf{x}_I$, where $\mathbf{x}_R$ and
    $\mathbf{x}_I$ are its real and imaginary parts (for the first layer
    $\mathbf{x}=\mathbf{Z}$, so $\mathbf{x}_R,\mathbf{x}_I$ are the stacked
    similarity tensors of Eq.~(\ref{eq:zpack}); for deeper layers they are the
    preceding layer's complex activation). The learnable kernel is itself
    complex, $\mathbf{W}=\mathbf{W}_R + i\,\mathbf{W}_I$, with
    $\mathbf{W}_R,\mathbf{W}_I$ two real-valued convolution kernels of
    identical shape. The complex convolution $\mathbf{W}\ast\mathbf{x}$
    then has real and imaginary parts~\cite{Trabelsi2018Deep}
    \begin{equation}
    \mathrm{Re}(\mathbf{W}\ast\mathbf{x}) = \mathbf{W}_R\ast\mathbf{x}_R - \mathbf{W}_I\ast\mathbf{x}_I,
    \quad
    \mathrm{Im}(\mathbf{W}\ast\mathbf{x}) = \mathbf{W}_R\ast\mathbf{x}_I + \mathbf{W}_I\ast\mathbf{x}_R,
    \end{equation}
    where $\ast$ denotes real 2-D convolution over the
    $(\text{time},\text{class})$ grid. This is followed by per-part batch
        normalisation and a complex ReLU~\cite{Trabelsi2018Deep}
        (Fig.~\ref{fig:complexcnn}). \\

   Unlike a real CNN, whose unconstrained $2C{\to}2C$ real mixing of the
      stacked $(\mathbf{x}_R,\mathbf{x}_I)$ would use four independent kernels
      $\bigl[\begin{smallmatrix}\mathbf{W}_{11} & \mathbf{W}_{12}\\
      \mathbf{W}_{21} & \mathbf{W}_{22}\end{smallmatrix}\bigr]$: the complex
      layer ties this block to
      $\bigl[\begin{smallmatrix}\mathbf{W}_R & -\mathbf{W}_I\\
      \mathbf{W}_I & \mathbf{W}_R\end{smallmatrix}\bigr]$, i.e.\ two free kernels
      reused (with a sign flip) across all four slots, forcing the real and
      imaginary halves to be mixed jointly at every layer. We stress that this is
      a \emph{constraint on how the two parts mix}, not a parameter saving in our
      setting: a complex layer with $C$ complex channels still carries twice the
      real weights of a real layer with $C$ channels. We therefore compare against
      a real baseline \emph{widened to the same total parameter count}
      (Tab.~\ref{tab:complex-vs-real}), isolating the effect of the structural
      prior from capacity.
 Layers 1--4 use
    temporal-only kernels of size $(k_t,1)=(5,1)$ that mix information
    across time steps but never across class indices; the final layer is a
    $1\times 1$ convolution that collapses the channel dimension.
    Restricting the kernels to be category-agnostic is essential under the
    open-vocabulary protocol: a kernel with non-trivial extent along the
    class axis would tie its weights to the specific class index ordering
    seen during training. We then read out the magnitude of the complex
    output,
    \begin{equation}
    \hat{\mathbf{Y}} = \big| \mathrm{ComplexCNN}(\mathbf{Z}) \big|
    \in \mathbb{R}^{T\times(C+1)},
    \end{equation}
    and apply a min--max rescale along the class axis to obtain the
    per-segment class scores $\hat{\mathbf{P}}\in\mathbb{R}^{T\times(C+1)}$.

\subsection{Training objective and trainable parameters}
\label{sec:method-train}

\textbf{Frozen modules}: ImageBind ($\mathcal{E}_a$, $\mathcal{E}_v$,
$\mathcal{E}_t$, $\sim 1.2$B params), and the CycleGAN generator
$G_{\mathrm{IF}\to\mathrm{mel}}$ ($\sim 11$M params, trained offline).

\textbf{Trainable modules}: The four temporal-attention modules
($\Phi_v$, $\Phi_v^{\textsc{i}}$, $\Phi_a$, $\Phi_\phi$) and the
ComplexCNN are the trainable modules in the architecture. The two audio--visual temporal attentions inside
$\mathcal{E}_v/\mathcal{E}_a$ together account for $\sim 4.2$M params;
the iHSV and CycleGAN-phase temporal attentions add $\sim 2.1$M
each. The ComplexCNN itself adds only $\sim 33$k params. 

\textbf{Loss}: Per-segment cross-entropy against the segment-level
label $y_t\in\mathcal{C}$:
\begin{equation}
\mathcal{L} = -\frac{1}{T}\sum_{t=1}^{T} \log\hat{\mathbf{P}}\big[t,y_t\big].
\end{equation}
At inference, $\hat{y}_t = \arg\max_c \hat{\mathbf{P}}[t,c]$. Unseen
classes are introduced at test time by extending only the text prompt
set $\mathbf{T}$; no parameters are updated and no per-class head
exists in the model.

{For the loss we change only the input to the loss-estimation code of~\cite{Zhou_2025_CVPR}: we take the magnitude of our complex-valued outputs, apply min--max normalisation, and pass the result to that code (PyTorch applies the softmax before computing the cross-entropy).}

\section{Experiments and Results}
\label{sec:results}

\subsection{Datasets and Splits}
\label{sec:dataset}

\textbf{OV-AVEBench~\cite{Zhou_2025_CVPR}} is an open-vocabulary AVEL benchmark
  with a fixed partition of \textbf{46 seen} and \textbf{21 unseen} categories plus a
  shared ``other'' (background) class. It has $24{,}800$ ten-second clips
  ($13{,}182$/$5{,}798$/$5{,}820$ train/val/test) at one frame per second, so each clip
  carries $T{=}10$ frame-level labels and at most one event category. We report three test
  partitions: \textbf{seen} ($1{,}664$ clips), \textbf{unseen} ($4{,}156$ clips), and
  \textbf{total} ($5{,}820$ clips).
\textbf{AVE-OV} re-casts the original AVE~\cite{tian2018avel} benchmark
    ($4{,}010$ clips, 28 categories) into the same open-vocabulary setting, assigning
    \textbf{20 seen} and \textbf{8 unseen} categories (plus ``other''). After dropping clips that do
    not fall under the assigned seen/unseen partition, the re-cast retains $3{,}675$
    clips: $2{,}945$/$329$/$401$, train/val/test with the same $T{=}10$ structure and
    seen/unseen/total reporting. Roughly $6\times$ smaller than OV-AVEBench and skewed toward human-action and instrument classes, it serves as an out-of-distribution
    stress test.
\vspace{-1em}
\subsection{Evaluation Metrics}
\label{sec:metrics}

We follow the official OV-AVEL evaluation~\cite{Zhou_2025_CVPR} and
report three complementary metrics, all computed against the per-frame
ground truth $y_t \in \{0,\dots,C-1,C\}$ where $0$ to $C-1$ are the indices of the foreground classes and $C$ is the background index.\\
\textbf{Per-frame accuracy (Acc).}
$\mathrm{Acc} = \frac{1}{N \cdot T}\sum_{n,t}
\mathbbm{1}[\hat y^{(n)}_t = y^{(n)}_t]$,
with $N$ clips of length $T{=}10$. Acc averages every frame uniformly.\\
\textbf{Segment-level F-score (Seg.):}
Per clip, frame-wise precision, recall, and $F_1$ over the union
$\{0,\dots,C\}$, macro-averaged over clips. Sensitive to the balance
between foreground recall and background precision.\\
\textbf{Event-level F-score (Eve.):}
A predicted event is the maximal contiguous run of frames sharing the
same non-background class, counted as a true positive if it overlaps
the corresponding ground-truth event with $\mathrm{IoU} \geq 0.5$
\emph{and} the predicted class matches. Macro-averaged over clips.
This metric penalizes boundary jitter, label flicker, and class
confusion simultaneously, and is the strictest.

\subsection{Implementation}
\label{sec:impl}

We build on frozen ImageBind-Huge encoders (ViT-H/14 vision, audio, text;
  $\sim$1.2\,B params, none updated)~\cite{girdhar2023imagebind}. Each of the four
  streams is refined by an identical single-layer Transformer self-attention block
  ($d_{\mathrm{model}}{=}1024$, 1 head, dropout $0.1$),
  and the four text-aligned similarity tensors are min--max normalised along the class
  axis and fused as two complex channels by a 5-layer ComplexCNN ($k_t{=}5$, CReLU,
  complex BN; $\sim$33\,k params). We train with per-frame cross-entropy, Adam at
  $\mathrm{lr}{=}5\!\times\!10^{-5}$, batch size $32$, selecting the best-validation
  checkpoint by $(\mathrm{Acc}{+}\mathrm{Seg}{+}\mathrm{Eve})/3$ on the validation
  total split.

{Inference cost, memory and latency for both variants against the fine-tuned baseline are reported in the supplementary material: the 2-stream adds negligible latency over the baseline, while the 4-stream adds roughly $+135$~ms.}

\vspace{-1em}
  \subsection{Experimental Results}
  \label{sec:main-results}

Table~\ref{tab:ovavel-main} compares our four-stream complex fusion against two
  baselines from the original OV-AVEL paper~\cite{Zhou_2025_CVPR}: a training-free
  baseline combining frozen audio-visual similarities via $\sqrt{\mathrm{ReLU}(s_v\cdot s_a)}$,
  and a fine-tuned baseline adding a real-valued temporal-attention head on the same
  frozen backbones (all numbers in \%). On the strictest metric, Event-level F1, our
  architecture beats all prior work by a substantial margin.
  To confirm the gains are not specific to OV-AVEBench, Table~\ref{tab:ave-ov-main}
  replicates the comparison on \textbf{AVE-OV}, a second open-vocabulary benchmark
  re-cast from AVE (Sec.~\ref{sec:dataset}), with all models trained and evaluated
  within AVE-OV under one identical protocol. The ranking holds: the 4-stream
  complex model is best on all twelve cells. Against the fine-tuned baseline it gains
  $+8.9$/$+5.9$/$+8.4$ Avg on seen/unseen/total, with the largest per-metric jump on
  the strict Event-F1 ($+12.6$ seen, $+9.6$ unseen). As on OV-AVEBench, the two extra
  streams help most on the unseen split ($+2.8$ Avg vs.\ $+0.3$ on seen), and the
  training-free baseline collapses on this smaller prompt set ($31.5$ total Avg):
  the learned complex fusion is what carries open-vocabulary performance. {For additional context, a comparison of our closed-set setup against the recent fully-supervised AVSGN (CVPR'25)~\cite{liu2025avsgn} is provided in the supplementary material; even restricted to the closed-set regime, our architecture is competitive with this dedicated closed-set method.}

  \begin{table*}[t]
  \centering
  \resizebox{\textwidth}{!}{%
  \begin{tabular}{|l|cccc|cccc|cccc|}
  \hline    
   & \multicolumn{4}{c|}{Seen} & \multicolumn{4}{c|}{Unseen} & \multicolumn{4}{c|}{Total} \\
  \hline
  Method & Acc & Seg & Eve & Avg & Acc & Seg & Eve & Avg & Acc & Seg & Eve & Avg \\
  \hline\hline
  CMRA~\cite{CMRAN} & 65.2 & 58.8 & 54.3 & 59.4 & 36.0 & 31.0 & 26.3 & 31.1 & 44.3 & 38.9 & 34.3 & 39.2 \\
  AVE~\cite{tian2018avel} & \textcolor{red}{\textbf{76.6}} & 63.6 & 56.0 & 65.4 & 44.6 & 33.2 & 24.0 & 34.0 & 53.8 & 41.9 & 33.2 & 42.9 \\
  PSP~\cite{PSP} & 75.4 & 66.8 & 61.0 & \textcolor{blue}{\textbf{67.7}} & 33.7 & 28.2 & 24.2 & 28.7 & 45.6 & 39.3 & 34.7 & 39.9 \\
  MM-Pyramid~\cite{MMPyramid} & \textcolor{blue}{\textbf{76.5}} & \textcolor{blue}{\textbf{66.9}} & \textcolor{blue}{\textbf{62.3}} & \textcolor{red}{\textbf{68.6}} & 36.8 & 29.0 &
  23.8 & 29.9 & 48.4 & 40.2 & 35.2 & 41.2 \\
  Baseline~\cite{Zhou_2025_CVPR}   & 72.5 & 61.8 & 54.5 & 62.9 & 64.9 & 55.0 & 47.5 & 55.8 & 67.1 & 56.9 & 49.5 & 57.8 \\
  \hline
  \textbf{Ours} (2-stream + ComplexCNN) & 72.8 & 65.1 & 61.0 & 66.3 & \textcolor{blue}{\textbf{65.3}} & \textcolor{blue}{\textbf{57.9}} & \textcolor{blue}{\textbf{53.3}} &
  \textcolor{blue}{\textbf{58.8}} & \textcolor{blue}{\textbf{67.4}} & \textcolor{blue}{\textbf{59.9}} & \textcolor{blue}{\textbf{55.5}} & \textcolor{blue}{\textbf{60.9}} \\
  \hline    
  \textbf{Ours} (4-stream + ComplexCNN) & 74.9 & \textcolor{red}{\textbf{67.4}} & \textcolor{red}{\textbf{63.6}} & \textcolor{red}{\textbf{68.6}} & \textcolor{red}{\textbf{66.5}} &
  \textcolor{red}{\textbf{59.1}} & \textcolor{red}{\textbf{54.1}} & \textcolor{red}{\textbf{59.9}} & \textcolor{red}{\textbf{68.2}} & \textcolor{red}{\textbf{61.1}} &
  \textcolor{red}{\textbf{57.0}} & \textcolor{red}{\textbf{62.1}} \\
  \hline
  \end{tabular}%
  }
  \vspace{2mm}
  \caption{\textbf{Results on OV-AVEBench} (\%). The \textcolor{red}{\textbf{best}} and \textcolor{blue}{\textbf{second best}} results are highlighted in red and blue, respectively.
  The finetune-baseline and the previous works are from~\cite{Zhou_2025_CVPR}. 
    }
  \label{tab:ovavel-main}
  \end{table*}

  \begin{table*}[t]
  \centering
  \resizebox{\textwidth}{!}{%
  \begin{tabular}{|l|cccc|cccc|cccc|}
  \hline    
   & \multicolumn{4}{c|}{Seen} & \multicolumn{4}{c|}{Unseen} & \multicolumn{4}{c|}{Total} \\
  \hline
  Method & Acc & Seg & Eve & Avg & Acc & Seg & Eve & Avg & Acc & Seg & Eve & Avg \\
  \hline\hline
  Training-free~\cite{Zhou_2025_CVPR}                     & 43.3 & 32.7 & 19.9 & 32.0 & 40.1 & 30.4 & 18.8 & 29.8 & 42.6 & 32.2 & 19.7 & 31.5 \\
  Fine-tuned baseline~\cite{Zhou_2025_CVPR}     & 82.3 & 74.6 & 70.5 & 75.8 & 58.0 & 46.7 & 40.8 & 48.5 & 77.2 & 68.4 & 63.9 & 69.8 \\
  \textbf{Ours} (2-stream complex)                        & 87.8 & 82.7 & 82.8 & 84.4 & 57.0 & 49.7 & 48.1 & 51.6 & 81.2 & 75.7 & 75.3 & 77.4 \\
  \textbf{Ours} (4-stream + ComplexCNN)                   & \textbf{88.0} & \textbf{82.9} & \textbf{83.1} & \textbf{84.7} & \textbf{60.7} & \textbf{51.9} & \textbf{50.4} &
  \textbf{54.4} & \textbf{82.2} & \textbf{76.4} & \textbf{76.1} & \textbf{78.2} \\
  \hline
  \end{tabular}%
  }
  \vspace{2mm}
  \caption{\textbf{Results on AVE-OV.} All values in \%.
  The same architecture wins on every split. On the unseen split, the
  4-stream complex fusion improves the re-implemented fine-tuned
  baseline by $+2.7/+5.2/+9.6/+5.9$ points (Acc/Seg/Eve/Avg), and
  improves the 2-stream complex variant by $+3.7/+2.2/+2.3/+2.8$.}
  \label{tab:ave-ov-main}
  \end{table*}

\begin{figure*}[t]
  \centering
  \includegraphics[width=\textwidth]{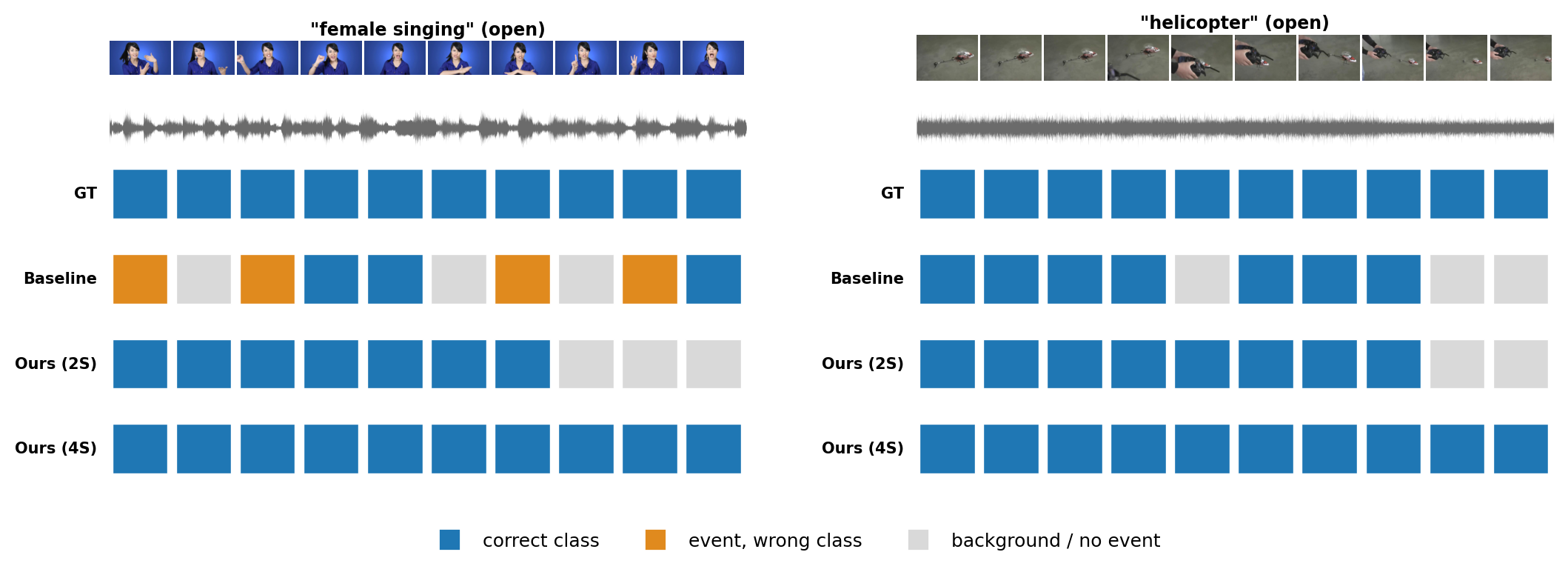}
  \caption{\textbf{Per-segment predictions on unseen-class videos.}
  Two examples side-by-side; each panel shows the 10 video frames, the
  time-aligned audio envelope, and four rows: ground-truth labels, the
  OV-AVEL fine-tuned baseline~\cite{Zhou_2025_CVPR} (geometric-mean
  fusion, trained temporal-attention heads), our 2-stream complex
  fusion, and our 4-stream complex fusion.}
\label{fig:timelines}
  \end{figure*}

Fig.~\ref{fig:timelines} shows qualitative per-segment predictions on
        two unseen-class OV-AVEBench clips for the
        baseline~\cite{Zhou_2025_CVPR}, 2-stream, and 4-stream fusion (blue =
        correct, orange = wrong class, grey = background). On ``female
        singing'' the baseline confuses the event with an acoustically
        similar class (orange) and drops several segments to background
        (grey); on ``helicopter'' it misses segments as background (grey).
        The 2-stream complex model recovers most segments and the 4-stream
        model is all-blue on both clips.

\vspace{-1em}
\subsubsection{Cross-dataset transfer}
    \label{sec:cross-dataset}
    Another type of open-vocabulary task could be cross-dataset transfer where we train the model on seen categories of one dataset and test on unseen categories of unseen dataset (text features are
    re-tokenised at inference; the class-agnostic fusion CNN needs no
    reshaping), reporting both directions in
    Tabs.~\ref{tab:cross-ov2ave}--\ref{tab:cross-ave2ov}. The 2-stream
    complex fusion keeps a positive margin over the fine-tuned baseline in
    both directions ($+3.6$/$+8.5$ Avg), while the 4-stream variant
    transfers unevenly: relative to the 2-stream variant it drops
    $-4.8$ Acc on OV$\rightarrow$AVE but gains $+2.1$ Acc on
    AVE$\rightarrow$OV.


\begin{table*}[t]
    \centering
    
    \begin{minipage}[t]{0.48\textwidth}
        \centering
        \resizebox{\linewidth}{!}{%
        \begin{tabular}{|l|cccc|}
        \hline
        Method & Acc & Seg & Eve & Avg \\
        \hline\hline
        Baseline~\cite{Zhou_2025_CVPR}     & 46.1 & 34.1 & 25.0 & 35.1 \\
        \textbf{Ours} (2-stream complex)   & \textbf{46.8} & \textbf{35.6} & 33.8 & \textbf{38.7} \\
        \textbf{Ours} (4-stream + ComplexCNN) & 42.0 & 34.4 & \textbf{34.3} & 36.9 \\
        \hline
        \end{tabular}%
        }
        \vspace{0.5mm}
        \caption{\textbf{OV-AVEBench $\rightarrow$ AVE-OV cross-dataset
        transfer} (\%). Each row is a checkpoint trained on the \emph{seen} categories of OV-AVEBench and
        evaluated zero-shot on the \emph{unseen} categories of the AVE-OV test split.}
        \label{tab:cross-ov2ave}
    \end{minipage}\hfill
    \begin{minipage}[t]{0.48\textwidth}
        \centering
        \resizebox{\linewidth}{!}{%
        \begin{tabular}{|l|cccc|}
        \hline
        Method & Acc & Seg & Eve & Avg \\
        \hline\hline
        Baseline~\cite{Zhou_2025_CVPR}     & 48.1 & 35.7 & 31.0 & 38.3 \\
        \textbf{Ours} (2-stream complex)   & 53.3 & 45.3 & 41.7 & 46.8 \\
        \textbf{Ours} (4-stream + ComplexCNN) & \textbf{55.4} & \textbf{47.4} & \textbf{43.8} & \textbf{48.9} \\
        \hline
        \end{tabular}%
        }
        \vspace{0.5mm}
        \caption{\textbf{AVE-OV $\rightarrow$ OV-AVEBench cross-dataset
        transfer} (\%). Each row is a checkpoint trained on the \emph{seen} categories of AVE-OV and
        evaluated zero-shot on the \emph{unseen} categories of the OV-AVEBench test split.}
        \label{tab:cross-ave2ov}
    \end{minipage}
    
\end{table*}
\subsection{Ablation Studies}
We conduct two ablations studies. One to compare the performance of our Complex-valued CNN against a parameter matched Real-valued CNN. The other is to study the effect of the frozen CycleGAN in our 4-stream architecture and how it effects our performance.

{We further ablate the real/imaginary channel assignment (swapping the two roles); the swap experiments confirm that our design choice (no swap) yields the best performance, and are reported in the supplementary material.}

\vspace{-1em}
\subsubsection{Ablation: ComplexCNN vs Real CNN fusion}
  \label{sec:ablation-real-vs-complex}
  
  To isolate the complex inductive bias, we compare ComplexCNN against a
  parameter-matched stacked RealCNN that consumes the same similarity tensors as
  real-valued channels (widened hidden channels match the parameter count). We vary
  one factor at a time: 2- vs.\ 4-stream input, Real vs.\ Complex fusion, under
  an identical training protocol. Tab.~\ref{tab:complex-vs-real} reports the unseen
  categories, where generalisation is most sensitive to architectural choices.
  
  \begin{table}[t]
  \centering
  \small
  \begin{minipage}[t]{0.49\linewidth}
  \centering
  {\footnotesize\textbf{(a) OV-AVEBench}}\\[2pt]
  \resizebox{\linewidth}{!}{%
  \begin{tabular}{|l|cccc|}
  \hline
  Configuration & Acc & Seg & Eve & Avg \\
  \hline\hline
  2-stream + RealCNN                    & 61.1 & 52.0 & 45.4 & 52.8 \\
  2-stream + \textbf{ComplexCNN}        & 65.3 & 57.9 & 53.3 & 58.8 \\
  4-stream + RealCNN                    & 64.3 & 56.7 & 52.1 & 57.7 \\
  \textbf{4-stream + ComplexCNN}        & \textbf{66.5} & \textbf{59.1} & \textbf{54.1} & \textbf{59.9} \\
  \hline
  \end{tabular}}
  \end{minipage}\hfill
  \begin{minipage}[t]{0.49\linewidth}
  \centering
  {\footnotesize\textbf{(b) AVE-OV (20/8)}}\\[2pt]
  \resizebox{\linewidth}{!}{%
  \begin{tabular}{|l|cccc|}
  \hline
  Configuration & Acc & Seg & Eve & Avg \\
  \hline\hline
  2-stream + RealCNN                    & 50.1 & 40.4 & 32.2 & 40.9 \\
  2-stream + \textbf{ComplexCNN}        & 57.0 & 49.7 & 48.1 & 51.6 \\
  4-stream + RealCNN                    & 58.5 & 51.2 & \textbf{50.4} & 53.4 \\
  \textbf{4-stream + ComplexCNN}        & \textbf{60.7} & \textbf{51.9} & \textbf{50.4} & \textbf{54.4} \\
  \hline
  \end{tabular}}
  \end{minipage}
  \vspace{2mm}
  \caption{\textbf{Complex vs.\ real fusion at matched parameters}
  (unseen categories). 
  }
  \label{tab:complex-vs-real}
  \end{table}

  Two findings.
  \textbf{(i) Complex parameterisation is the dominant lever.}
  Holding streams constant, real$\to$complex gives $+7.9$ Event-F1 at
  2-stream and $+2.0$ at 4-stream on OV-AVEBench; on AVE-OV the gap is
  $+15.9$ at 2-stream and a tie at 4-stream Event-F1, with the complex
  variant still ahead on Acc/Avg.
  \textbf{(ii) Auxiliary streams add a smaller, complementary gain:
  only the complex fusion exploits them.} On top of the 2-stream complex
  model, adding iHSV + CycleGAN-phase lifts unseen Event-F1 by $+0.8$ on
  OV-AVEBench and $+2.3$ on AVE-OV. Under real fusion the same streams give
  a larger raw jump ($+6.7$ / $+18.2$ Event-F1) because the real baseline
  is so weak, yet on OV-AVEBench the 4-stream real model still trails the
  2-stream complex one, confirming the two ingredients are complementary
  rather than substitutes.
  \subsubsection{Ablation: CycleGAN-translated phase vs raw IF}
    \label{sec:ablation-cyclegan}

    The audio imaginary stream (Sec.~\ref{sec:method-audio-imag}) uses an
    unpaired CycleGAN to translate the instantaneous-frequency
    spectrogram into the mel manifold before passing it to the frozen
    audio encoder. To check whether this translation step matters, we
    replace it with the raw IF spectrogram
    and otherwise keep the architecture identical. Both runs share
    hyperparameters and the 4-stream complex-pair fusion; only the
    audio imaginary stream changes. Table~\ref{tab:phase-ablation} shows the results of how CycleGAN helps in the OV-AVEL task.
    Feeding the raw IF directly drops Event-F1 by $2.2$/$1.9$ points on
    OV-AVEBench/AVE-OV, so the gain comes from the translation step, not from
    merely adding a phase-derived input: the frozen encoder, pretrained on
    mel statistics, cannot interpret off-manifold IF.
    Notably, the drop is largest on the strictest metric (Event-F1) while
    Accuracy barely moves and AVE-OV Seg-F1 is unchanged, indicating that
    the translated phase mainly sharpens event boundaries and class
    consistency rather than per-frame labelling. This is consistent with
    our design goal: the CycleGAN does not add new information so much as
    re-express the phase cues in the magnitude-mel domain the frozen
    encoder was trained on, so the otherwise-discarded onset and transient
    structure becomes usable. {Note that the CycleGAN inherently preserves the audio phase information in the generated mel-spectrogram; otherwise it would not be able to reconstruct the phase back (the cycle-consistency loss).}

\begin{table}[t]
\centering
\begin{minipage}[t]{0.485\linewidth}
\centering
{\footnotesize\textbf{(a) OV-AVEBench}}\\[2pt]
\resizebox{\linewidth}{!}{%
\begin{tabular}{|l|ccc|}
\hline
Audio imag.\ stream & Acc & Seg & Eve \\
\hline\hline
Raw IF              & 64.7 & 56.7 & 51.9 \\
CycleGAN      & \textbf{66.5} & \textbf{59.1} & \textbf{54.1} \\
\hline
$\Delta$            & +1.8 & \textbf{+2.4} & \textbf{+2.2} \\
\hline
\end{tabular}}
\end{minipage}\hfill
\begin{minipage}[t]{0.485\linewidth}
\centering
{\footnotesize\textbf{(b) AVE-OV (20/8)}}\\[2pt]
\resizebox{\linewidth}{!}{%
\begin{tabular}{|l|ccc|}
\hline
Audio imag.\ stream & Acc & Seg & Eve \\
\hline\hline
Raw IF              & 60.0 & \textbf{51.9} & 48.5 \\
CycleGAN     & \textbf{60.7} & \textbf{51.9} & \textbf{50.4} \\
\hline
$\Delta$            & +0.7 & +0.0 & \textbf{+1.9} \\
\hline
\end{tabular}}
\end{minipage}
\vspace{2mm}
\caption{\textbf{CycleGAN-translated phase outperforms raw IF}
  (unseen categories, 4-stream complex fusion; only the audio
  imaginary stream changes). CycleGAN improves every metric on
  \textbf{(a) OV-AVEBench} ($+1.8/+2.4/+2.2$ Acc/Seg/Eve) and
  Event-F1 on \textbf{(b) AVE-OV} ($+1.9$; Seg-F1 tied).
  }
\label{tab:phase-ablation}
\end{table}
\FloatBarrier
{A backbone-generality check with a frozen LanguageBind backbone, where the same ranking holds, is provided in the supplementary material.}

\vspace{-1em}
\section{Conclusion}
\label{sec:conclusion}
  We propose two layered architectural contributions for OV-AVEL.
  \emph{First}, a complex-valued fusion that processes the visual--text
  and audio--text similarities as the real and imaginary parts of a
  single complex value; the fusion is a five-layer Complex-valued CNN
  whose complex weights tie the cross-channel mixing into a structured
  block rather than an unconstrained real map --- a constraint we
  evaluate against a parameter-matched real baseline. This 2-stream
  complex fusion alone produces the dominant gain over the previous
  work's scalar geometric-mean fusion. \emph{Second}, a four-stream
  extension that enriches each modality with a paired companion stream
  (iHSV for vision, CycleGAN-translated phase for audio) and feeds the
  resulting complex content--structure pairs to the same complexcnn;
  the auxiliary streams add a smaller and complementary gain on top of
  (1). On OV-AVEBench the full architecture improves the fine-tuned
  baseline by $+6.6$ Event-F1 on the unseen split ($+1.6/+4.1$ Acc/Seg);
  seen and total partitions move in the same direction ($+9.1$ and
  $+7.5$ Event-F1). The two contributions are isolated by separate
  ablations: the complex-vs-real breakdown (Tab.~\ref{tab:complex-vs-real})
  shows complex fusion is the dominant lever (real$\to$complex at 2
  streams: $+7.9$ Event-F1 OV-AVEBench, $+15.9$ AVE-OV) and the auxiliary
  streams add a smaller additive boost (adding iHSV + CycleGAN-phase to
  the 2-stream complex model: $+0.8$ / $+2.3$ Event-F1); at matched
  parameters complex over real fusion at 4 streams adds $+2.2$ Avg on
  OV-AVEBench and $+1.0$ on AVE-OV (unseen); and CycleGAN-translated
  phase adds $+2.2$ Event-F1 over raw IF (Tab.~\ref{tab:phase-ablation}). {Our 4-stream approach outperformed the 2-stream approach on both datasets. Only in one of the two cross-dataset transfer experiments, our 4-stream approach performed worse than 2-stream. This could be due to its complicated nature, as it might have over-fitted on the larger dataset it was trained on and ended
up exhibiting lesser cross-dataset transfer capabilities.}

\bibliography{egbib}
\end{document}


\maketitle

\noindent
This supplementary material accompanies the main paper and provides four
additional analyses: (\textbf{\S\ref{sec:supp-swap}}) a real/imaginary
channel-assignment ablation; (\textbf{\S\ref{sec:supp-closedset}}) a contextual
comparison against a recent fully-supervised closed-set method;
(\textbf{\S\ref{sec:supp-cost}}) the inference cost, memory and latency of both
variants; and (\textbf{\S\ref{sec:supp-backbone}}) a backbone-generality check
with LanguageBind. 

\section{Real/imaginary channel-assignment ablation}
\label{sec:supp-swap}

The main paper assigns the visual--text similarity to the \emph{real} part and
the audio--text similarity to the \emph{imaginary} part of the 2-stream complex
input, and, in the 4-stream model, the RGB/mel content views to the real parts
and the iHSV/CycleGAN-phase companions to the imaginary parts. To verify that
this assignment is load-bearing rather than arbitrary, we re-train with the
real and imaginary roles swapped, one channel at a time, holding every other
component: streams, fusion module, training protocol --- identical.
Table~\ref{tab:supp-swap} reports the unseen (open-vocabulary) split of
OV-AVEBench, the setting where generalisation is most sensitive to
architectural choices.

\begin{table}[t]
\centering
\resizebox{0.75\textwidth}{!}{%
\begin{tabular}{|l|cccc|}
\hline
Real$\leftrightarrow$Imag assignment (Unseen) & Acc & Seg & Eve & Avg \\
\hline\hline
\multicolumn{5}{|l|}{\emph{2-stream, $z=\mathrm{sim}_{vt}+i\,\mathrm{sim}_{at}$ (ours)}} \\
\hline
\quad no swap (ours)          & 65.3 & \textbf{57.9} & \textbf{53.3} & \textbf{58.8} \\
\quad swapped                 & \textbf{65.4} & 57.6 & 52.0 & 58.3 \\
\hline\hline
\multicolumn{5}{|l|}{\emph{4-stream (ours)}} \\
\hline
\quad no swap (ours)          & \textbf{66.5} & \textbf{59.1} & \textbf{54.1} & \textbf{59.9} \\
\quad visual channel swapped  & 66.2 & 58.5 & 53.9 & 59.5 \\
\quad audio channel swapped   & 65.5 & 57.2 & 51.6 & 58.1 \\
\quad both swapped            & 65.0 & 57.2 & 52.5 & 58.2 \\
\hline
\end{tabular}%
}
\vspace{2mm}
\caption{\textbf{Real/imaginary channel-assignment ablation} (unseen
categories, OV-AVEBench). Real and imaginary roles are swapped one channel at a
time; all other components are held fixed. The main-paper assignment
(``no swap'') is best on Avg and on Event-F1 in both variants, and the audio
channel is the most sensitive to swapping. All values in \%.}
\label{tab:supp-swap}
\end{table}

Two observations.
\textbf{(i) The chosen assignment is best on the summary metrics.} In both the
2- and 4-stream models, ``no swap'' --- the assignment used throughout the main
paper which gives the best Avg and the best Event-F1, the strictest metric. In
the 2-stream model the swap is essentially tied on per-frame Acc ($-0.1$) yet
loses $1.3$ Event-F1, indicating the effect is on event/boundary consistency
rather than raw frame labelling.
\textbf{(ii) The audio channel is the more sensitive one.} In the 4-stream
model, swapping the audio real/imaginary roles (mel $\leftrightarrow$
CycleGAN-phase) costs $2.5$ Event-F1 more than swapping the visual roles
($-0.2$) or swapping both ($-1.6$). This matches the intuition that the
magnitude-trained encoder expects mel magnitude as the content (real) channel
and the translated phase as the structural (imaginary) companion; reversing
them feeds the frozen encoder the less canonical input as content.

\section{Contextual comparison with a fully-supervised closed-set method}
\label{sec:supp-closedset}

The main paper positions our method against the open-vocabulary state of the
art~\cite{Zhou_2025_CVPR}. For additional context we also report how the same
complex-fusion architecture behaves in a purely \emph{closed-set}, fully
supervised regime on the original AVE benchmark~\cite{tian2018avel}, where
train and test share the same category set and no open-vocabulary
generalisation is required. We compare against AVSGN~\cite{liu2025avsgn}, a
recent (CVPR'25) fully-supervised closed-set AVEL method. This is a sanity
check in a different setting rather than a like-for-like comparison: AVSGN is
designed for the closed-set task and does not perform open-vocabulary
localisation, whereas our architecture targets the strictly harder
open-vocabulary setting and is merely restricted here to seen classes.

\begin{table}[t]
\centering
\resizebox{0.6\textwidth}{!}{%
\begin{tabular}{|l|c|}
\hline
Method (seen classes / AVE) & Acc \\
\hline\hline
\multicolumn{2}{|l|}{\emph{Closed-set (fully supervised)}} \\
\hline
AVSGN (CVPR'25)~\cite{liu2025avsgn} & 83.1 \\
\hline\hline
\multicolumn{2}{|l|}{\emph{Our closed-set setup}} \\
\hline
Ours (2-stream + ComplexCNN) & 87.8 \\
Ours (4-stream + ComplexCNN) & \textbf{88.0} \\
\hline
\end{tabular}%
}
\vspace{2mm}
\caption{\textbf{Contextual comparison with a fully-supervised closed-set
method} on the AVE dataset (seen classes only, per-frame Acc). Our
complex-fusion architecture, restricted to the closed-set regime it was not
designed for, is competitive with a recent dedicated closed-set method despite
targeting the harder open-vocabulary task. All values in \%.}
\label{tab:supp-closedset}
\end{table}

Even restricted to the closed-set regime, our architecture exceeds the
fully-supervised AVSGN baseline by $+4.7$/$+4.9$ Acc (2-/4-stream). We stress
that this is a contextual sanity check: the paper's contribution is the
open-vocabulary fusion of the main results, not a new closed-set AVEL method.

\section{Inference cost, memory and latency}
\label{sec:supp-cost}

Table~\ref{tab:supp-cost} reports the inference-time cost of both variants
against the fine-tuned baseline~\cite{Zhou_2025_CVPR}, measured on a single
Nvidia RTX Pro 6000 for one 10-second clip ($T{=}10$ segments) and averaged over the
test set. All three configurations share the same frozen ImageBind backbone;
the reported trainable-parameter counts exclude the frozen encoders.

\begin{table}[t]
\centering
\resizebox{0.7\textwidth}{!}{%
\begin{tabular}{|l|ccc|}
\hline
Inference, one clip & Baseline & 2-stream & 4-stream \\
\hline\hline
Latency (ms)         & 127.4 & 128.1 & 263.2 \\
FLOPs (G)            & 1817  & 1817  & 3829 \\
Memory (GB)          & 4.77  & 4.77  & 4.82 \\
Trainable params (M) & 8.42  & 8.45  & 12.66 \\
\hline
\end{tabular}%
}
\vspace{2mm}
\caption{\textbf{Inference cost, memory and latency} of the 2- and 4-stream
variants against the fine-tuned baseline~\cite{Zhou_2025_CVPR}, one 10-second
clip. The 2-stream fusion is almost free over the baseline; the 4-stream fusion
roughly doubles latency and FLOPs for its additional gain.}
\label{tab:supp-cost}
\end{table}

\textbf{The 2-stream variant is nearly free.} It adds $0.7$ ms of latency and
$0.03$M trainable parameters over the baseline (identical FLOPs and memory to
two significant figures) while delivering the bulk of the accuracy gain
reported in the main paper. \textbf{The 4-stream variant roughly doubles
latency} ($+135$ ms over the baseline) and FLOPs, because the two auxiliary
streams each require an additional forward pass through the frozen encoder, and
adds $\sim 4.2$M trainable parameters; memory is essentially unchanged
($+0.05$ GB) since the frozen backbone dominates the footprint. The 4-stream
model is thus the accuracy-oriented configuration and the 2-stream model the
efficiency-oriented one.

\section{Backbone generality: LanguageBind}
\label{sec:supp-backbone}

The main paper uses a frozen ImageBind~\cite{girdhar2023imagebind} backbone. To
check that the complex-fusion gains are not specific to that backbone, we repeat
the comparison with a frozen LanguageBind~\cite{LanguageBind} backbone, keeping
the rest of the pipeline unchanged. Table~\ref{tab:supp-lb} reports the unseen
(open-vocabulary) split of OV-AVEBench.

\begin{table}[t]
\centering
\resizebox{0.85\linewidth}{!}{%
\begin{tabular}{|l|cccc|}
\hline
LanguageBind (unseen)              & Acc & Seg & Eve & Avg \\
\hline\hline
geomean baseline (no fusion head)  & 64.62 & 54.10 & 44.87 & 54.53 \\
2-stream + \textbf{ComplexCNN}     & 64.64 & 55.78 & 50.40 & 56.92 \\
4-stream + \textbf{ComplexCNN}     & \textbf{64.85} & \textbf{56.31} & \textbf{51.50} & \textbf{57.55} \\
\hline
\end{tabular}%
}
\vspace{2mm}
\caption{\textbf{Backbone generality} (unseen categories, OV-AVEBench,
LanguageBind backbone). The same pattern holds as with ImageBind: both complex
variants beat the geometric-mean baseline, and the 4-stream variant is best. All
values in \%.}
\label{tab:supp-lb}
\end{table}
\FloatBarrier
The ranking is unchanged from the ImageBind results in the main paper: both
complex variants improve over the geometric-mean baseline, and the 4-stream
variant is best on every metric ($+3.0$ Avg over the baseline, $+0.6$ over the
2-stream variant). The complex-fusion gain therefore transfers across
backbones and is not an artifact of the specific ImageBind features.

\bibliography{egbib}